\documentclass[journal]{IEEEtran}
\usepackage{amsmath,amsfonts}
\usepackage{array}
\usepackage[caption=false,font=normalsize,labelfont=sf,textfont=sf]{subfig}
\usepackage{textcomp}
\usepackage{stfloats}
\usepackage{url}
\usepackage{verbatim}
\usepackage{graphicx}
\usepackage{cite}
\usepackage{orcidlink}
\usepackage{multirow}
\usepackage[linesnumbered,ruled,vlined]{algorithm2e}

\begin{document}

\title{OV-VG: A Benchmark for Open-Vocabulary Visual Grounding}

\author{Chunlei Wang\orcidlink{0000-0002-8955-9964}, Wenquan Feng, Xiangtai Li\orcidlink{0000-0002-0550-8247},~\IEEEmembership{Member,~IEEE}, Guangliang Cheng\orcidlink{0000-0001-8686-9513}, Shuchang Lyu\orcidlink{0000-0001-9769-7083},~\IEEEmembership{Graduate Student Member,~IEEE,} Binghao Liu\orcidlink{0000-0001-6590-0016}, Lijiang Chen\orcidlink{0000-0001-7732-3527}, Qi Zhao\orcidlink{0000-0002-3508-027X},~\IEEEmembership{Member,~IEEE}
\thanks{This work was supported in part by the National Natural Science Foundation of China under Grants 62072021.(\emph{Corresponding author: Qi Zhao and Xiangtai Li})}
\thanks{Chunlei Wang, Wenquan Feng, Shuchang Lyu, Binghao Liu, Lijiang Chen, Qi Zhao are with the Department of Electronics and Information Engineering, Beihang University, Beijing 100191, China(e-mail:\{wcl\_buaa; buaafwq; lyushuchang; liubinghao; chenlijiang; zhaoqi\}@buaa.edu.cn).}
\thanks{Xiangtai Li is with the S-Lab, Nanyang Technological University, Singapore. (e-mail:xiangtai.li@ntu.edu.sg).}
\thanks{Guangliang Cheng is with the Department of Computer Science, University of Liverpool. (e-mail:Guangliang.Cheng@liverpool.ac.uk).}
}

\markboth{IEEE Transactions on Circuits and Systems for Video Technology}
{Shell \MakeLowercase{\textit{et al.}}: A Sample Article Using IEEEtran.cls for IEEE Journals}


\maketitle
\begin{abstract}
    \par Open-vocabulary learning has emerged as a cutting-edge research area, particularly in light of the widespread adoption of vision-based foundational models. Its primary objective is to comprehend novel concepts that are not encompassed within a predefined vocabulary. One key facet of this endeavor is Visual Grounding (VG), which entails locating a specific region within an image based on a corresponding language description. While current foundational models excel at various visual language tasks, there's a noticeable absence of models specifically tailored for open-vocabulary visual grounding (OV-VG). This research endeavor introduces novel and challenging OV tasks, namely Open-Vocabulary Visual Grounding (OV-VG) and Open-Vocabulary Phrase Localization (OV-PL). The overarching aim is to establish connections between language descriptions and the localization of novel objects. To facilitate this, we have curated a comprehensive annotated benchmark, encompassing 7,272 OV-VG images (comprising 10,000 instances) and 1,000 OV-PL images. In our pursuit of addressing these challenges, we delved into various baseline methodologies rooted in existing open-vocabulary object detection (OV-D), VG, and phrase localization (PL) frameworks. Surprisingly, we discovered that state-of-the-art (SOTA) methods often falter in diverse scenarios. Consequently, we developed a novel framework that integrates two critical components: Text-Image Query Selection (TIQS) and Language-Guided Feature Attention (LGFA). These modules are designed to bolster the recognition of novel categories and enhance the alignment between visual and linguistic information. Extensive experiments demonstrate the efficacy of our proposed framework, which consistently attains SOTA performance across the OV-VG task. Additionally, ablation studies provide further evidence of the effectiveness of our innovative models. Codes and datasets will be made publicly available at \url{https://github.com/cv516Buaa/OV-VG}.
\end{abstract}


\section{Introduction}
\label{intro}
\IEEEPARstart{V}{isual} grounding (VG) revolves around the objective of precisely locating target objects within an image based on linguistic references. It serves as a cornerstone in computer vision, facilitating enhanced understanding of visual-linguistic interactions and closing the semantic gap, which holds immense potential for practical applications, including but not limited to robot navigation~\cite{robot_navigation} and visual dialog~\cite{visual_dialog}. While previous approaches~\cite{Yangetal-vltvg, Dengetal-TransVG} have made notable advancements in enhancing visual-linguistic alignment by investigating feature representations that bridge the gap between vision and language, they fall short in the crucial task of detecting novel objects, which is a challenging and practical problem in applications. To the best of our knowledge, no publicly available datasets have been designed specifically to support the detection of novel categories solely relying on base-category annotations in the context of visual grounding tasks. 
\begin{figure}[t]
\begin{center}
   \includegraphics[width=1.0\linewidth]{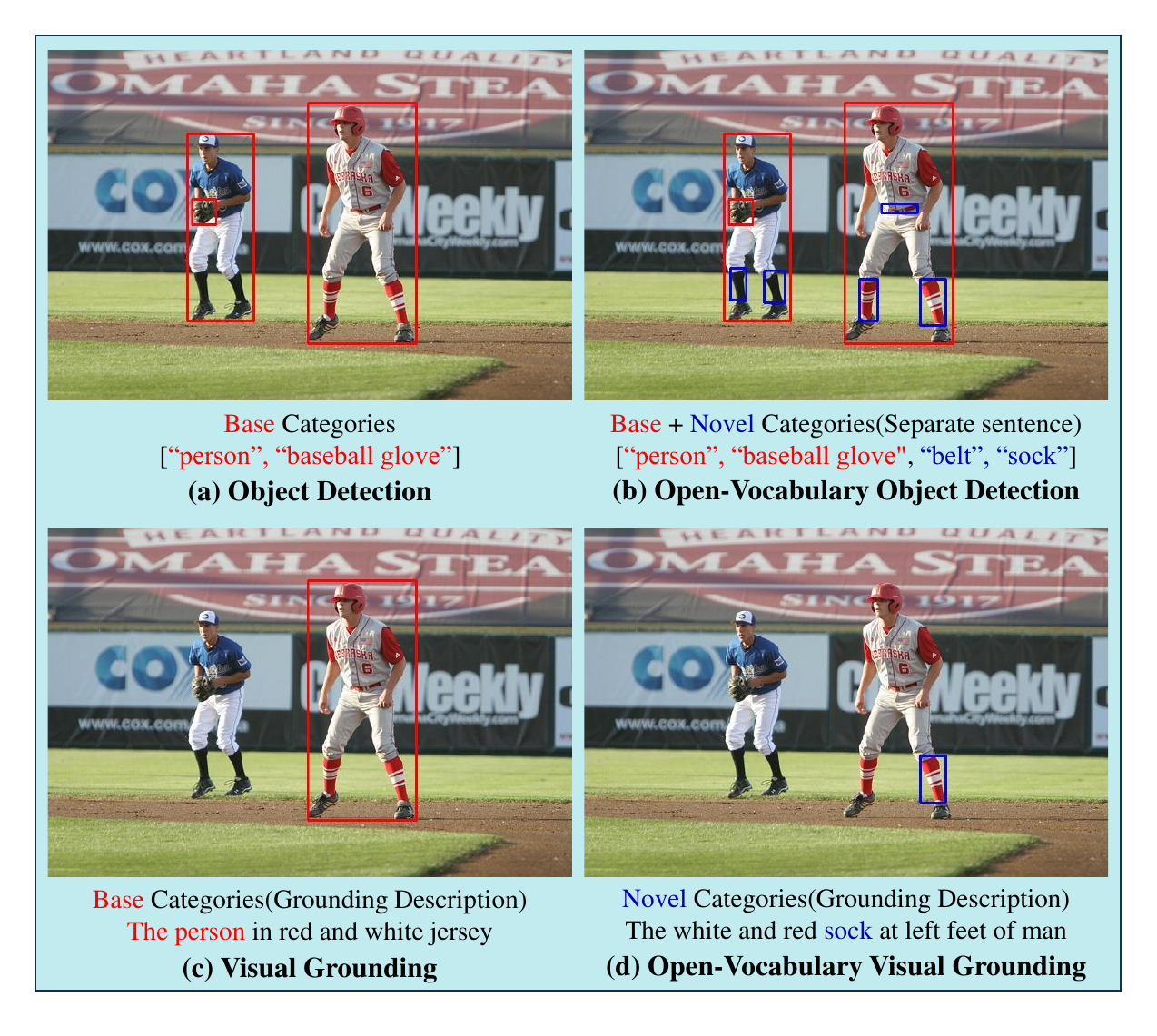}
\end{center}
   \caption{Different task settings. (a) Traditional object detection. (b) Open-vocabulary object detection. (c) Traditional visual grounding. (d) Our proposed open-vocabulary visual grounding.}
\label{problem_setting}
\end{figure}
\par Recently, open-vocabulary learning has garnered significant attention within the research community. It addresses the formidable challenge of enhancing perceptual capabilities to recognize novel categories with the guidance of natural language. Recent developments such as CLIP~\cite{Radfordetal-CLIP} and foundation models~\cite{liuetal-groundingDINO, lietal-GLIP, li2023panopticpartformer++, li2023transformer, fang2023explore} have spurred a wave of research into open-vocabulary detection (OV-D) and open-vocabulary segmentation (OV-S)~\cite{kirillovetal-SAM}, which aims to enable the identification of novel objects, entirely reliant on the base-category annotations. However, existing open-vocabulary algorithms suffer from data leakage, which means the model has been trained on a large amount of data, leading to the occurrence of novel categories during training. Data leakage can indeed improve model performance on novel categories, while it's not the strictly zero-shot or open-vocabulary definition.

\par To address the above issues, this paper introduces a challenging benchmark dataset tailor-made for the open-vocabulary visual grounding (OV-VG) task.
We present an innovative network architecture designed for this specific task. Specifically, we design and release an OV-VG benchmark dataset comprising 100 novel categories, each with about 100 instances, totaling 10,000 instances. Our OV-VG dataset poses numerous challenges, such as handling extensive and detailed object descriptions, managing substantial disparities in object sizes, and accommodating diverse object categories. Our innovative approach, which incorporates text-image query selection (TIQS) and language-guided feature attention (LGFA) techniques, excels in improving the alignment between visuals and language and the comprehension of global semantic context across the entire image. 
\par

\par
\begin{table}[tbp]
\renewcommand{\arraystretch}{1.5}
\begin{center}
\caption{Differences between OV-VG, OV-PL and existing tasks}
\label{problem_setting_tab}
\begin{tabular}{c|cc|cc}
\hline
& \begin{tabular}[c]{@{}c@{}}long\\ sentence\end{tabular} & \begin{tabular}[c]{@{}c@{}}open\\ vocabulary\end{tabular} & \begin{tabular}[c]{@{}c@{}}Multiple\\ instances\end{tabular} & \begin{tabular}[c]{@{}c@{}}specific\\ target\end{tabular} \\ \hline
VG & $\surd$ & ${\rm{ \times }}$ & ${\rm{ \times }}$ & $\surd$ \\
PL & $\surd$ & ${\rm{ \times }}$ & $\surd$ & ${\rm{ \times }}$ \\
OV-D & × & $\surd$ & $\surd$ & ${\rm{ \times }}$ \\
OV-PL & $\surd$ & $\surd$ & $\surd$ & ${\rm{ \times }}$ \\
OV-VG & $\surd$ & $\surd$ & ${\rm{ \times }}$ & $\surd$ \\ \hline
\end{tabular}
\end{center}
\end{table}
At the same time, we introduce the first open-vocabulary phrase localization (OV-PL) dataset, consisting of 1000 images. In this dataset, each image is accompanied by two descriptions: one exclusively encompasses basic categories, while the other incorporates a combination of basic and novel categories. Additionally, we provide several baseline models tailored to the OV-PL dataset. Furthermore, we differentiate our OV-VG and OV-PL task configurations from VG, PL, and OV-D, as summarized in Table~\ref{problem_setting_tab}.
\par The main contributions are summarized as follows:
\begin{itemize}
  \item [$\bullet$]
  We propose open-vocabulary visual grounding (OV-VG) and open-vocabulary phrase localization (OV-PL) problem settings and release two benchmark datasets for further research.\par
  \item [$\bullet$]
  We benchmark the proposed OV-VG and OV-PL datasets built upon existing methods.\par
  \item [$\bullet$]
  We design an effective network that incorporates text-image query selection (TIQS) and language-guided feature attention (LGFA) for open-vocabulary visual grounding to enhance the recognition of novel categories and strengthen visual-linguistic understanding.
  \item [$\bullet$]
  Extensive experiments demonstrate the effectiveness of our proposed method on the OV-VG dataset, whether in settings involving data leakage or not.
\end{itemize}
\par The main research content of this paper is outlined as follows: In Sec.~\ref{related_work}, we introduce related work on visual grounding, phrase localization, and some open-vocabulary-based algorithms. In Sec.~\ref{dataset}, we provide a detailed explanation of our dataset construction. In Sec.~\ref{method}, we describe our method and network details. In Sec.~\ref{experiments}, we design extensive experiments, and ablation studies are conducted to verify the effectiveness of the proposed method. Finally, we conclude this paper in Sec.~\ref{conclusion}. 

\section{Related Work} 
\label{related_work}
\subsection{Visual Grounding} 
Visual grounding is a critical task that involves providing a precise target-object location within an image based on a corresponding natural language description. Within the realm of visual grounding, existing methods can be categorized into two groups: two-stage methods~\cite{Huetal-CMNs,Liuetal-NMTree,Hongetal-LTSV} and one-stage methods~\cite{Chenetal-SSG,TCSVT_RES,wu2022towards, TMM_one_stage}. Most existing visual grounding frameworks are the extension of object detection methods \cite{Huetal-CMNs, Dengetal-TransVG}.
\par In two-stage approaches, the initial step involves generating region proposals, followed by leveraging specific language input to identify the most suitable proposal. Prior research has explored a combination of tree structures~\cite{Liuetal-NMTree, Hongetal-LTSV} and modular designs~\cite{Huetal-CMNs} to derive region scores. However, two-stage methods have faced criticism for their relatively slow inference speeds. On the other hand, one-stage approaches seamlessly integrate visual and language features to pinpoint the specific region of interest directly. While renowned for their simplicity and efficiency, one-stage methods face a challenge in capturing a holistic, contextual understanding from the fusion of vision and language information due to the limitations of pointwise feature representation.
\par However, whether it involves region proposals or dense anchor boxes, identifying target objects with very detailed language descriptions can be challenging. Transformer develops rapidly in computer vision, leading to the emergence of transformer-based visual grounding~\cite{Dengetal-TransVG,Yangetal-vltvg,IR-VG}, which enables direct retrieval of target features for localization. TransVG~\cite{Dengetal-TransVG} initially formulates visual grounding as a direct coordinate regression task and introduces visual-linguistic fusion modules that use self-attention to embed input tokens from both intra-modality and inter-modality into a common semantic space. VLTVG~\cite{Yangetal-vltvg} introduces visual-language verification to construct discriminative feature maps and employs context aggregation to gather the contextual features, making the visual features of the target object more distinguishable. 
\subsection{Phrase Localization}
\par Phrase localization seeks to establish associations between noun phrases and specific regions within images. Traditionally, researchers have differentiated between entities and image regions by introducing spatial relationships within phrase-image pairs~\cite{Wangetal-SMPL, Plummer-Cues4PL}. However, in recent years, the advent of transformer-based models~\cite{Zhaoetal-MATN, Liuetal-LCMCG, Liuetal-Visualbert} has ushered in a new era in phrase localization. These models have empowered the extraction of both textual and visual context information, offering exciting prospects for advancing this field.
Nonetheless, this task is confronted with formidable challenges due to the expensive ground-truth annotations and the inherent susceptibility to human error. Consequently, weakly supervised~\cite{Zhaoetal-MATN, Datta-Align2ground, TMM_PL} and unsupervised~\cite{Wangetal-PLpaired} methodologies have progressively gained prominence in the realm of phrase localization. Align2Ground~\cite{Datta-Align2ground} leverages caption-to-image retrieval as a ``downstream'' task to guide the phrase localization process. This paper introduces a novel open-vocabulary phrase localization benchmark and presents multiple baseline approaches employing the latest state-of-the-art models. 

\subsection{Open-Vocabulary Learning}
Open-vocabulary learning seeks to broaden vocabulary and comprehension. Previous works~\cite{cheng2020panoptic,xiangtl_decouple,Li2020SRNet,sfnet,htc,cai2018cascade,wang2020solov2,li2022videoknet,li2023tube,wang2021max} for scene understanding follow the close-set assumption and try to maximize the performance for limited label space.
Its successful application spans diverse domains, encompassing tasks such as object detection \cite{OVD-kd1, OVD-kd2, OVD-kd3}, instance segmentation~\cite{openseed, freeseg, X-decoder}, video comprehension~\cite{Pointclip, Clip2point}, and various visual language challenges~\cite{wu2023towards}.
The mainstream open-vocabulary object detection (OV-D) can be divided into five categories: 1) Knowledge distillation~\cite{guetal-ViLD, OVD-kd1, OVD-kd2, OVD-kd3, TCSVT_KD,clipself} aims to distill the knowledge of Visual-Language Models into close-set detectors. 2) Region text pre-training~\cite{Zareianetal-OVRCNN, lietal-GLIP, zhangetal-GLIPv2, TCSVT_TI, TCSVT_TR} aims to map the visual features and text embeddings into the same feature space. 3) training with more balanced data~\cite{mmovod, OVD-balance1, OVD-balance2, dst_ov_det} leverage more balanced classification datasets with pseudo labels to joint training the models. 4) prompting modeling~\cite{ov-detr, cora, prompt-ovd, detpro} generates text embeddings of category names, and prompts are fed to the text encoder of pre-trained VLMs. 5) Region text alignment~\cite{kuoetal-fvlm, yaoetal-detclip, guetal-ViLD, detclipv2} uses language as supervision instead of ground-truth bounding boxes. For instance, ViLD~\cite{guetal-ViLD} distills text embedding and image embedding from pre-trained open-vocabulary models for training two-stage open-vocabulary detectors. F-VLM~\cite{kuoetal-fvlm} freezes the vision language models and finetunes only the detector head to simplify open-vocabulary object detection. Grounding DINO~\cite{liuetal-groundingDINO} concatenates all category names as input text and outputs the highest scores for object detection.\par
Open-vocabulary segmentation (OV-S) encompasses several distinct technical approaches. Visual Language Models (VLMs) have demonstrated strong performance by learning to interpret visual language expressions for classification tasks, thereby facilitating transfer to OV-S~\cite{Fusioner, han_FOVS, OPSNet, tagclip}. Another avenue in OV-S is acquiring new class information through category names provided by classification data~\cite{OVseg, XPM, openseg, cgg}. Recognizing that segmentation entails multiple objectives, a noteworthy direction involves the simultaneous training of semantic segmentation and instance segmentation~\cite{openseed, freeseg, X-decoder, pomp}. Additionally, there has been a growing interest in leveraging diffusion models, as their intermediate representations often exhibit alignment with natural language vocabularies. This has led to the emergence of diffusion model-based OV-S methods~\cite{ODISE, OVDiff, diffumask, diffuguiding, xie2023mosaicfusion}. Notably, OpenSeeD~\cite{openseed} introduces a decoupled decoding model that seamlessly integrates segmentation and detection tasks, enabling the joint implementation of OV-D and OV-S. Similarly, X-Decoder~\cite{X-decoder} adopts a joint training approach for segmentation and image-text pairs, harnessing OV-S capabilities for downstream tasks. Furthermore, open vocabulary video comprehension and open vocabulary 3D comprehension have seen significant advancements in recent times~\cite{Pointclip, Clip2point}.

\subsection{Open-Vocabulary Visual Grounding} 
Open-vocabulary object detection approaches~\cite{OVD-kd1, OVD-kd2, OVD-kd3, openseed} aim to identify objects in images without relying on predefined object categories, allowing for a more flexible and adaptive recognition process. These methods can accept input as natural language phrases or extract relevant phrases from sentences, enabling them to detect a wide range of object categories in diverse contexts. They have gained significant attention in computer vision research due to their potential to handle novel and context-specific objects effectively. Unlike the OV-D task, OV-VG aims to enhance visual-linguistic understanding while identifying novel targeted category objects described in the long sentence by comprehending the relationships among instances. 

To the best of our knowledge, no existing benchmarks or approaches have been specifically tailored for the exploration of the open-vocabulary visual grounding task. Current visual grounding methods, such as VLTVG~\cite{Yangetal-vltvg}, encounter challenges when dealing with the open-vocabulary problem. The existing models built upon the OV-D framework primarily focus on object detection, implying that they invariably attempt to predict all objects within an image~\cite{TMM_T2I}. Given that language descriptions often do not precisely align with specific image regions, accurately identifying the target object can be a formidable task.

In this paper, we introduce two benchmark datasets: OV-VG and OV-PL. We provide a range of baseline models for OV-VG tasks, grounded in both VG and OV-D frameworks. Furthermore, we introduce several phrase localization methods within our OV-PL dataset. Lastly, we bridge the gap between OV-D and VG methodologies, proposing a novel network to address the challenges posed by the OV-VG problem.

\section{Dataset Construction} 
  \label{dataset}
In this section, we will introduce OV-VG and OV-PL datasets in detail, including dataset description, category selection, and labeling strategy. We also analyze these two datasets and give some examples to illustrate their characteristics. 
\subsection{Dataset Descriptions}
The OV-VG dataset contains 7,272 images with 10,000 instances for open-vocabulary visual grounding. All of the images are selected from MS COCO~\cite{linetal-coco} and are \textbf{disjoint} with RefCOCO~\cite{yuetal-refcoco}, RefCOCO+~\cite{yuetal-refcoco} and RefCOCOg~\cite{maoetal-refcocog} training set. We choose 80 categories in COCO as base classes and 100 more common and suitable categories (disjoint with COCO) from LVIS~\cite{guptaetal-lvis} as novel classes. The novel categories encompass various aspects of the real world and ensure multiple novel instances in each image, which is essential for the requirements of the VG task. Furthermore, we have curated a set of 1,000 images from the OV-VG dataset to facilitate the open-vocabulary phrase localization task. These selected images encompass a diverse range of both base and novel instances. The annotation format is identical to Flickr30k Entities~\cite{plummeretal-flickr30kE}. In the following section, we will delve into the specifics of our dataset.

\begin{figure}[t]
\begin{center}
   \includegraphics[width=1.0\linewidth]{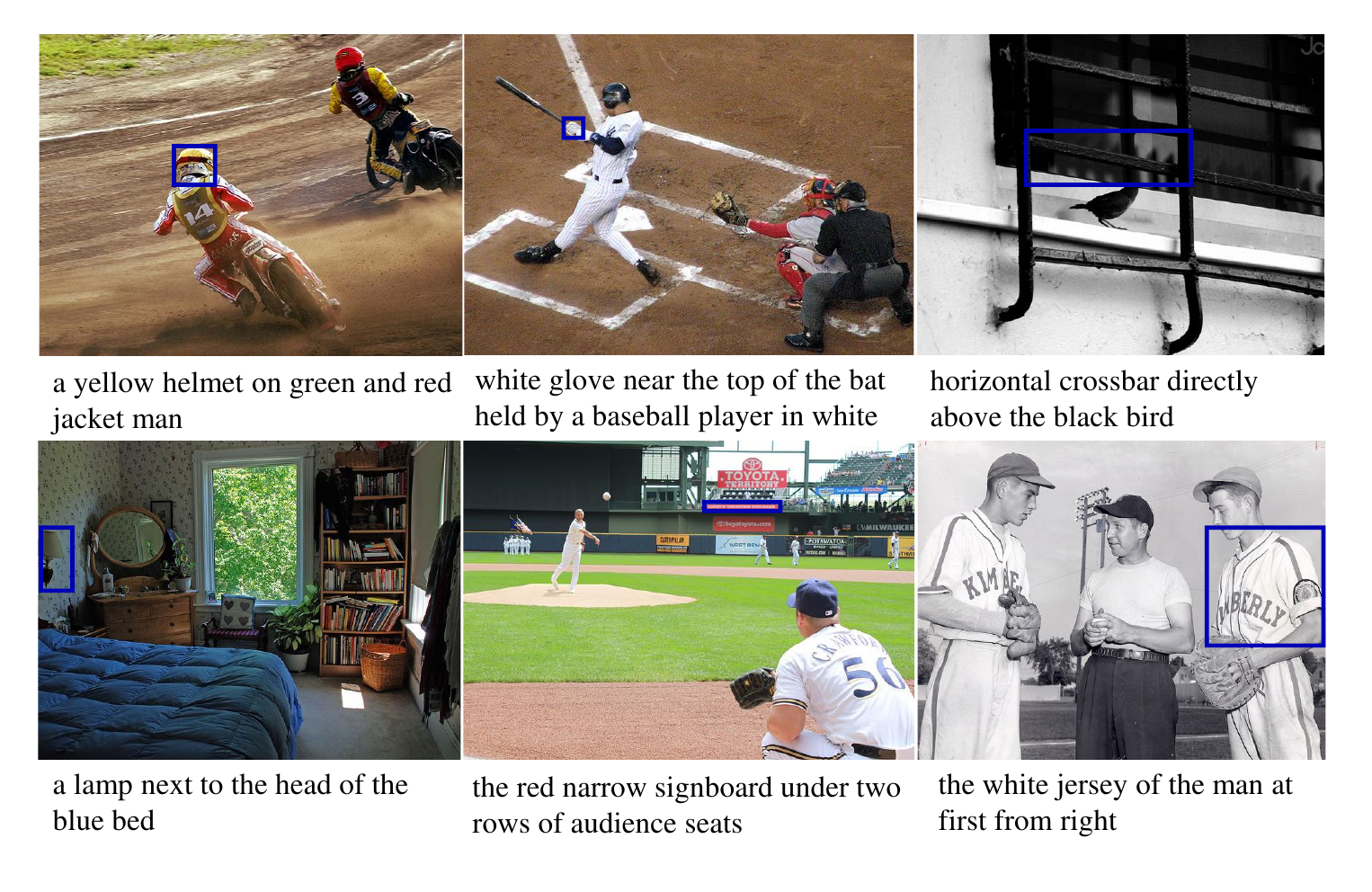}
\end{center}
   \caption{Samples of our OV-VG dataset. Blue boxes are the ground truths.}
\label{OVVG_samples}
\end{figure}

\subsection{Data Disjoint}
 \subsubsection{Image Disjoint}
 In the process of data annotation, we must ensure the independence of OV-VG relative to the training set. Since the training and testing sets of RefCOCO~\cite{yuetal-refcoco} and COCO2017~\cite{linetal-coco} intersect with each other, we select OV-VG images from the intersection of COCO2017 val and RefCOCO val. Therefore, it can be guaranteed that the images in the OV-VG and RefCOCO training set are disjoint. Although images in Flickr30k Entities and COCO are completely orthogonal, to enrich our dataset and task, we ensure that the images of OV-PL are completely from OV-VG. Since the phrase localization task requires as many instances as possible, we select 1000 images with the richest instances from the OV-VG dataset to construct our OV-PL dataset.
 \subsubsection{Category Disjoint}
 To ensure the category disjointness of novel and base categories. We have selected 80 categories from COCO as the base classes and 100 novel categories (disjoint with COCO) from LVIS~\cite{guptaetal-lvis}. LVIS contains more than 1000 categories, and these categories exhibit a long-tail distribution. To expand the dataset for subsequent studies, we have attempted to select novel categories with more instances. Considering that one of the challenges in the visual grounding task is to distinguish different instances of the same category within the input image, we have chosen images that contain multiple instances of the same novel category.
\begin{figure}[t]
\begin{center}
   \includegraphics[width=1.0\linewidth]{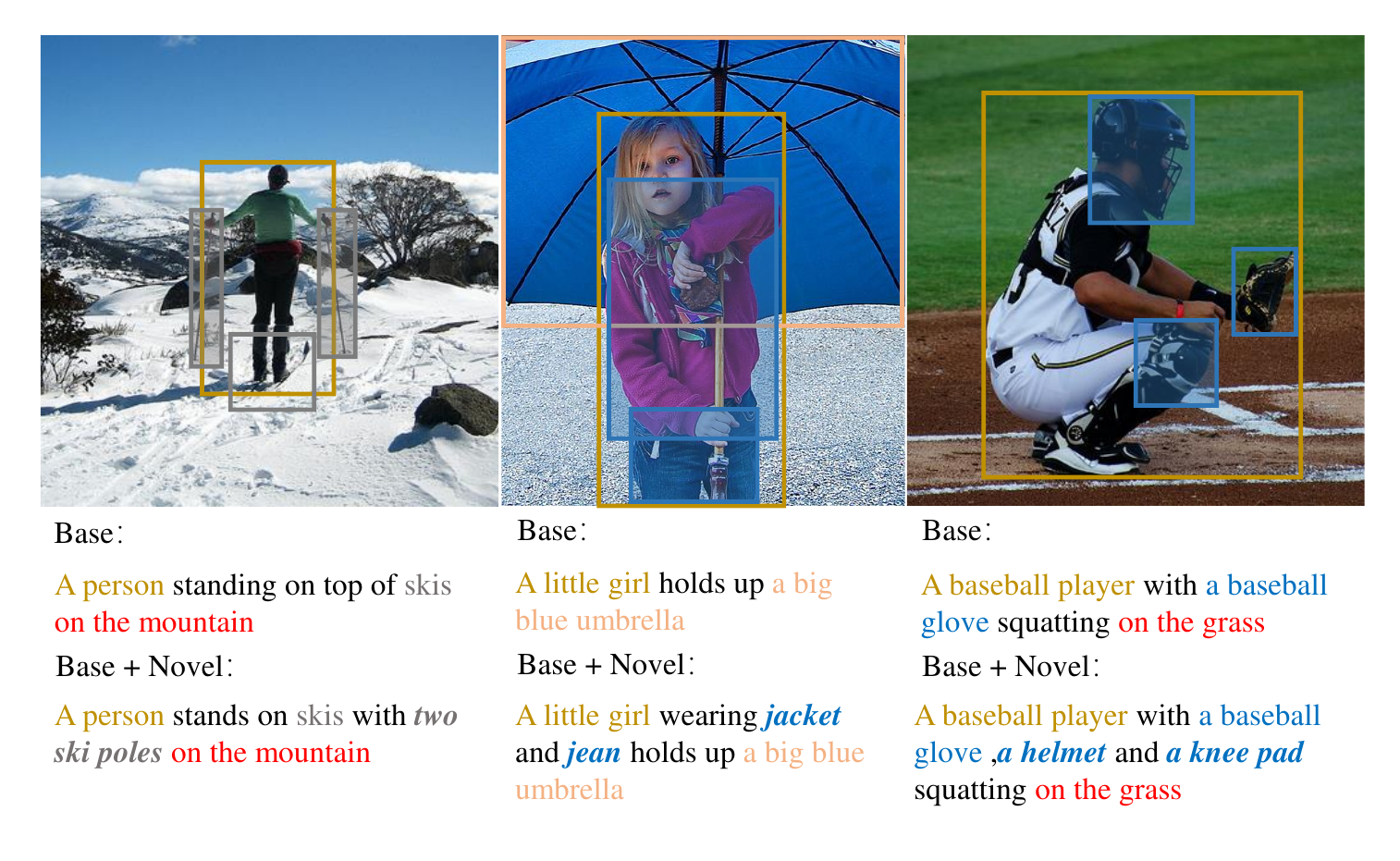}
\end{center}
   \caption{Samples of our OV-PL dataset. Each group of captions describes the same image. Coreferent mentions and their corresponding bounding boxes are marked with the same color. Bold and italics indicate novel categories.}
\label{OVPL_samples}
\end{figure}
\subsection{Data Annotation and Samples}
 \subsubsection{OV-VG Referring Expression Annotation}
\begin{figure}[t]
\begin{center}
   \includegraphics[width=1.0\linewidth]{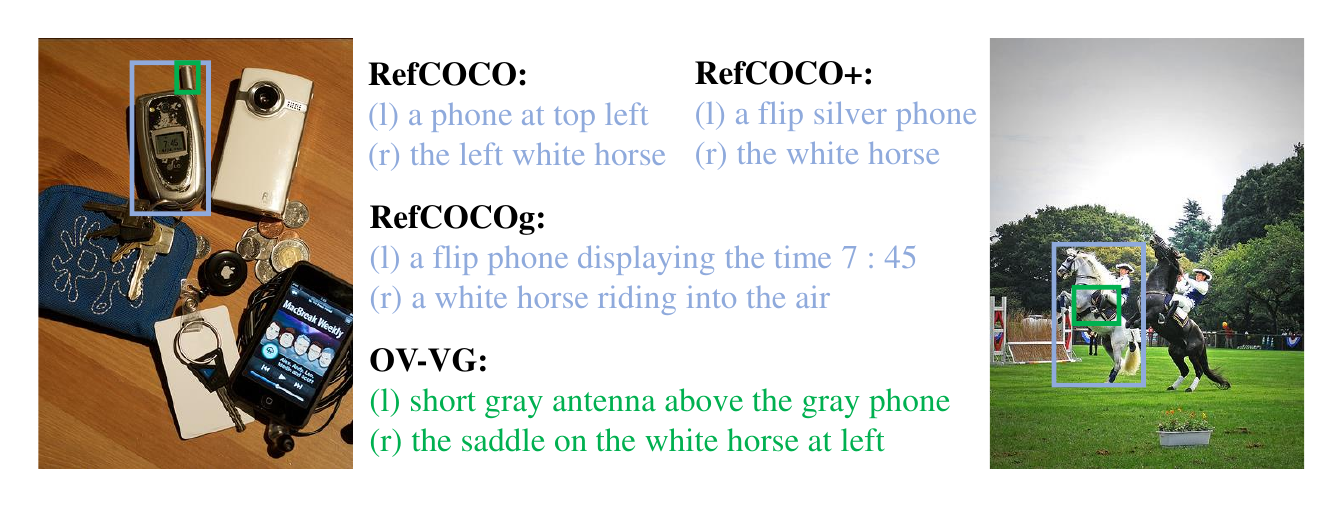}
\end{center}
   \caption{Examples of referring expression for existing VG and OV-VG datasets.}
\label{coco_ovvg}
\end{figure}

 We initiate the process by extracting object detection annotations specifically for novel objects from the LVIS dataset. To ensure the utmost accuracy and reliability of these annotations, we engaged a team of 6 annotation experts. Additionally, we enlisted the services of two quality checkers who meticulously double-checked the annotations for consistency and precision. Our annotation process focuses on capturing comprehensive descriptive information about each object, guided by its context within the image. Examples of OV-VG annotations are shown in Fig.~\ref{OVVG_samples}. We place the novel category representing at the beginning of each description, and our descriptions are exceptionally rich, including attributes (such as color and shape) and relative relations between objects within the same perceptual group (such as orientation and relationship among objects). The OV-VG dataset not only includes novel categories, but in the annotation process, we deliberately refine the description of the target object. Compared to existing visual grounding datasets~\cite{yuetal-refcoco,maoetal-refcocog}, the OV-VG dataset further enhances the focus on visual-linguistic understanding, which is the central aspect of VG. When comparing the annotation with existing visual grounding datasets, as shown in Fig.~\ref{coco_ovvg}, RefCOCO contains the target object and several position words, RefCOCO+ replaces absolute locations with action behaviors, and RefCOCOg uses more detailed descriptions. Our OV-VG descriptions resemble the form of RefCOCOg while aiming to describe the relationships of novel target objects in detail, without restricting the use of orientation and attribute descriptions. The average lengths of descriptions in RefCOCO, RefCOCO+, RefCOCOg, and OV-VG are 3.61, 3.53, 8.43, and 9.32, respectively.
 \subsubsection{OV-VG Bounding Box Annotation}
\begin{figure}[t]
\begin{center}
   \includegraphics[width=1.0\linewidth]{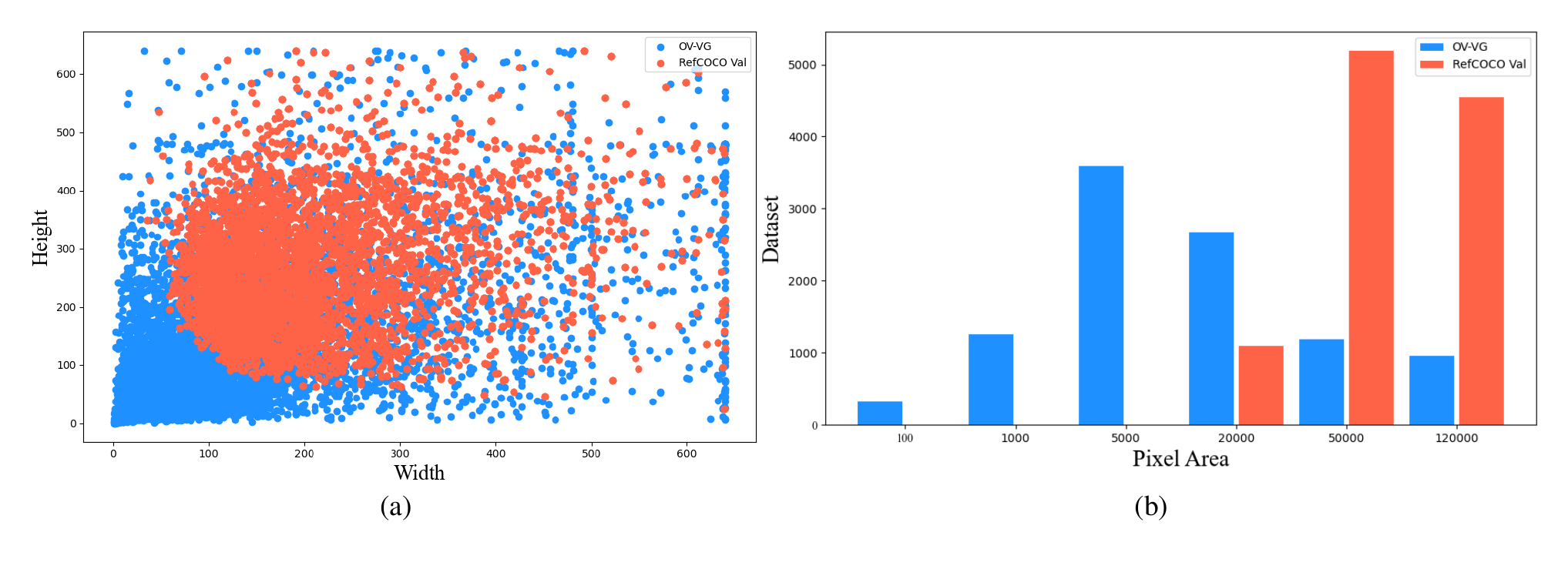}
\end{center}
   \caption{Data distribution of OV-VG and RefCOCO val. (a) Width and height distribution of the bounding box. (b) Statistics of the bounding box area. Blue is the bounding box annotation in OV-VG and the orange box is from RefCOCO val.}
\label{data_dis}
\end{figure}

 We utilize the annotation boxes from the original LVIS~\cite{guptaetal-lvis} dataset for the target objects as bounding boxes and process them into the same format as RefCOCO. These bounding boxes encode location and size as 4-dimensional vectors, representing the $x$ and $y$ locations of the top-left and bottom-right corners of the target object. It is worth noting that the target object bounding box presents more challenges in our OV-VG dataset. To enhance the precision of novel target object localization in response to the referring expression, our bounding boxes exhibit variable sizes. We compared the size of the target box in OV-VG and RefCOCO val, as shown in Fig.~\ref{data_dis}. The number of instances in OV-VG and RefCOCO val is almost the same, with 10,000 and 10,834 instances, respectively. However, real-world target objects are not as ideal as those in RefCOCO, where they are often large and nearly square. Compared with RefCOCO, the scale of target object annotation in our OV-VG dataset varies greatly, aligning more with real-world open-vocabulary situations. As shown in Fig.~\ref{data_dis} (a), the scatter points have a wider spread, indicating that OV-VG includes more objects with large aspect ratios than RefCOCO val. In Fig.~\ref{data_dis} (b), a significant number of targets are smaller than those in regular VG datasets but still within the typical object detection scale, and there are even some extremely small targets. This requires the network not only to align visual-linguistic information but also to accurately locate novel category targets, making it significantly more challenging. 
 \subsubsection{OV-PL Annotation}
 To improve the quality and uniformity of our dataset, we select 1000 images from OV-VG to constitute our OV-PL dataset. Our OV-PL annotations follow a structure similar to Flickr30k Entities~\cite{plummeretal-flickr30kE}, using the same highly structured format for overall annotation. However, due to the difficulty of annotation, and to distinguish the PL and OV-PL tasks, we provide only two structured description sentences for each image (compared to five sentences for each image in Flickr30k Entities). One sentence uses only base categories, while the other uses both base and novel categories. Our annotation pipeline consists of two stages: coreference resolution and bounding box annotation. It is worth mentioning that the entity mentioned in Flickr30k Entities rely on Flickr30k~\cite{Young-flickr30k}. In our case, we refer to the caption descriptions of images in COCO Caption~\cite{coco_caption} and LVIS\cite{guptaetal-lvis}. This results in different bounding box annotations compared to Flickr30k Entities. We export box annotations for both base and novel categories from COCO Caption and LVIS. We then describe the image based on the existing boxes and refer them to the entities. Following the rules of entity selection for open vocabulary, we manually annotate the scene descriptively. Specifically, we assume that any noun phrase (NP) chunk is a potential entity mention, which may refer to a single entity, multiple distinct entities, and groups of entities. Some surrounding NP chunks may not refer to any physical entities. Once we obtain the image caption, we need to identify which one refers to the same set of entities. We also collect binary coreference links between pairs of mentions as ~\cite{plummeretal-flickr30kE}. At this point, the phrase localization annotation for a single image is completed. We also need to unify and verify phrase references between images using a coreference chain verification task, following the same settings as~\cite{plummeretal-flickr30kE}.\par
 The OV-PL annotation examples are shown in Fig.~\ref{OVPL_samples}. Each image has two different descriptive annotations: one is described using only base category entities, and the other uses both base and novel categories. Translucent filled boxes in Fig.~\ref{OVPL_samples}, and the bold and italic phrases, represent novel categories. The same coreference chains are marked with the same color, e.g., golden represents all types of 'people' and blue represents all 'clothing'. Note that red expresses scenes and events ('on the mountain' and 'on the grass'), which have no boxes. In the left example, the 'two ski poles' chains point to multiple boxes. In the middle and right examples, each chain points to a single entity. 
\section{Methodology} 
 \label{method}
\subsection{The Overall Network}
\begin{figure*}[t]
\begin{center}
   \includegraphics[width=1.0\linewidth]{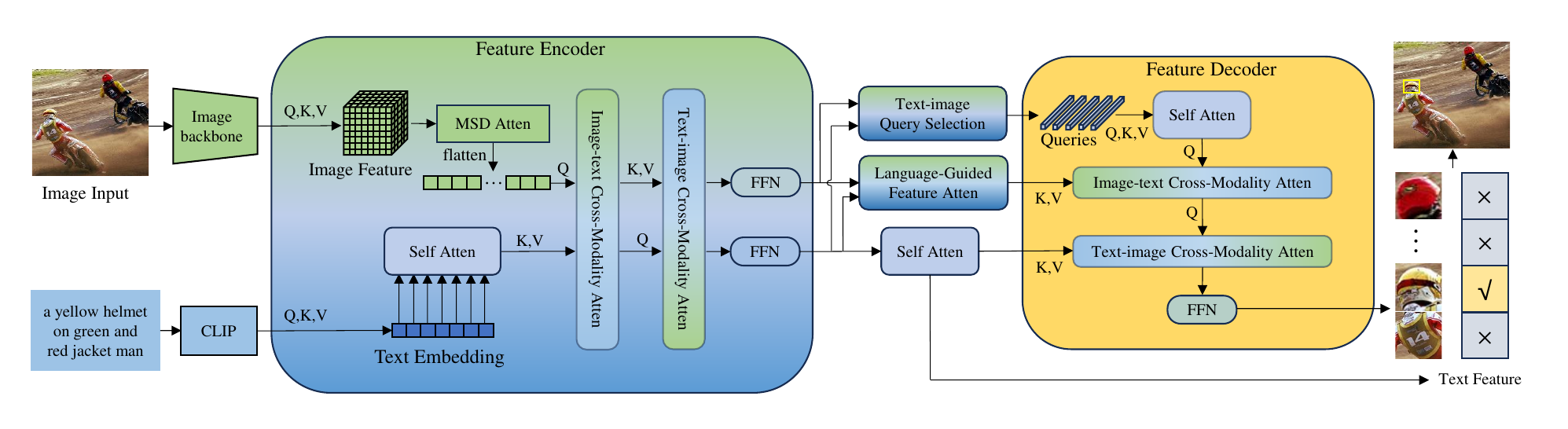}
\end{center}
   \caption{Overview of the proposed network, which comprises the encoder and decoder structure. The network consists of an LGFA module to guide the image activation area, and a TIQS module to combine text embedding with query selection and select the top-k boxes. MSD Atten stands for Multi-scale Deformable self-attention and FFN denotes feed-forward network.}
\label{method_overview}
\end{figure*}

  In this section, we present the detailed framework of our proposed method, as shown in Fig.~\ref{method_overview}. We combine the current popular VG and OV-D network structures to design our end-to-end OV-VG network. Our OV-VG network directly extracts the target object feature for localization. As shown in Fig.~\ref{method_overview}, given an (image, text) pair, we first extract the image features with an image backbone like ResNet-50~\cite{resnet} or Swin transformer~\cite{Swin_transformer}, and textual embedding with a text backbone like BERT~\cite{bert} or CLIP~\cite{Radfordetal-CLIP}. After that, we feed the image and text features into a feature encoder for feature fusion. To align these two modalities of features, inspired by GLIP~\cite{lietal-GLIP}, we add image-text and text-image cross-modality attentions in the feature encoder. Then, we apply language-guided feature attention (LGFA) and text-image query selection (TIQS) to further refer to the target object. The LGFA enforces the image features to focus on referring expression regions, while the TIQS provides all potential linguistically related localization boxes and selects top-k queries. Finally, the feature decoder is applied to analyze the encoded image and text features to more accurately localize the target object and output top-1 box. The pipeline process of our method is shown in Algorithm ~\ref{algorithm}.
\begin{algorithm}
    \SetAlgoLined
 \caption{Pipeline of our method}
        \label{algorithm}
 \KwIn{input image-text pair $I$ and $L$}
 \KwOut{the bounding box of target object}
 \textbf{Backbone:} output image feature $F_v$ and text embedding $F_l$ \par
 \textbf{Encoder:} after MSD attention and flatten image feature, obtain $v_v$ and after self attention we get text embedding $v_l$\par
 \For{$i = 0$ to $n$}{
 fuse image2text(${v_{v(i)}}$, $v_{l(i)}$);\par
 fuse text2image(${v_{v(i)}}$, ${v_{l(i)}}$);\par}
 output $v'_v$ and $v'_l$\par
 \textbf{TIQS:} calculate cosine similarity of ${v'_{v(i)}}$ and ${v'_{l(i)}}$ and select top-k queries\par
 \textbf{LGFA:} compute scores $S\left( x \right)$ of ${v'_{v(i)}}$ and ${v'_{l(i)}}$ according to formula (1) and dot product of ${v'_{v(i)}}$ and $S\left( {x} \right)$\par
 \textbf{Decoder:} after self-attention, we obtain the top-k queries \par
 \For{$i = 0$ to $n$}{
 fuse image2text(${v'_{v(i)}}$, ${v'_{l(i)}}$);\par
 fuse text2image(${v'_{v(i)}}$, ${v'_{l(i)}}$);\par}
 where $n$ is min(len(${v'_{v(i)}}$ layers), len(${v'_{l(i)}}$ layers)), we choose the goal of top-1 box as the target object
\end{algorithm}

\subsection{Feature Encoder}
Given an image and a language expression, we input them into the CLIP image and text backbone to extract the image feature and text embedding, respectively. We use multi-scale deformable self-attention to enhance and flatten image features, self-attention is used to enhance text features. Finally, we introduce two cross-modality attentions to deeply fuse the image and text information. In particular, we traverse and fuse each layer output of flattened image feature and text embedding. After fusing the current layer output, we update the input for the next layer to further fuse visual and language modalities. We define $n = min{\rm{ }}({N_v},{\rm{ }}{N_l})$, where${{N_v}}$ and ${{N_l}}$ are encoder layers of image and text, ${v'_v}$ can be shown as:
\begin{equation}
    {v'_v} = concat\left( {{{\rm{{\cal P}}}_{v \to l}}\left( {{{\rm{{\cal P}}}_{l \to v}}\left( {{v_{v(i)}}},{{v_{l(i)}}} \right)} \right)} \right),0 \le i \le n)
\end{equation}
${{\rm{{\cal P}}}_{v \to l}}\left(  \cdot  \right)$ and ${{\rm{{\cal P}}}_{l \to v}}\left(  \cdot  \right)$ mean image-to-text and text-to-image fusion, respectively. ${v_{v(i)}}$ and ${v_{l(i)}}$ are $i$-th layer of image and text encoder, and FFN is formed of two linear projection layers with ReLU activations. 
\subsection{Modular Structure}
In this subsection, we introduce the language-guided feature attention (LGFA) and text-image query selection (TIQS) of our network. We will cover the internal structure of these modules and explain the motivation.
\subsubsection{Language-Guided Feature Attention Model}
The language-guided feature attention model is based on multi-head attention, the query is flattened image feature ${v'_v}$, and the key and value are the text embedding ${v'_l}$. The multi-head attention aligns flatten image features with text embedding to generate semantic map ${v'_s}$, then we use linear projection and L2 normalization for mapping ${v'_v}$ and ${v'_s}$ to the same space, express as ${\hat v'_v}$ and ${\hat v'_s}$, then we calculate the attention score for each point ${x}$, denote $\alpha$ and $\sigma$ are learnable parameters:
\begin{equation}
    {S_x} = \alpha  \cdot \exp \left( { - \frac{{{{\left( {1 - {{\hat v'_v}}{{\left( {x} \right)}^{\bf{T}}}{\hat v'_s}\left( {x} \right)} \right)}^2}}}{{2{\sigma ^2}}}} \right)
\end{equation}
After obtaining language-guided image feature attention scores about semantic relevance, we can mathematically get the most relevant area in the image according to text information. Finally, we take the dot product result of the above-mentioned score and ${v'_v}$ as a new ${v''_v}$ to feature decoder, shown as
\begin{equation}
    {v''_v} = \beta  \cdot {v'_v} \cdot {S_x} + \left( {1 - \beta } \right) \cdot {v'_v}
\end{equation}
Where $\beta$ is a balance parameter, we empirically set ${\beta} = 0.7$.

\subsubsection{Text-image Query Selection}
To further improve visual-language understanding, we introduce a text-image query selection model. We first generate proposals and compute the einsum as the logits according to flatten image feature ${v'_v}$ and text embedding ${v'_l}$:
\begin{equation}
    {S_l} = \frac{{{v^{'}}_v^{\rm{T}}{v'_l}}}{{\left\| {{v'_v}} \right\|\left\| {{v'_l}} \right\|}}
\end{equation}
${S_l}$ denotes the logit scores. This function is used for the similarity measure between the flattened image feature ${v'_v}$ and the text embedding ${v'_l}$, with the aim of matching these two modality features. After that, we sort the proposals according to logit scores and select the top-k queries. The text-image query selection outputs k queries to the decoder query selection, with each decoder query selection including dynamic anchor boxes and content queries.
\subsection{Feature Decoder}
In order to select and localize the bounding box of the top-1 target object from the visual and language features, we improve the DINO decoder by adding several text attentions to align the text and image modalities better.\par Firstly, we take the top-k queries of text-image query selection as input to the feature decoder. The query ${t_q} \in \mathbb{R}{^{C \times 1}}$
is input to the self-attention model to collect semantic information of the referred object ${t_l} \in \mathbb{R}{^{C \times 1}}$, which acts as query, and the output of language-guided feature attention model acts as the key and value for the image-text cross-modality attention. In this manner, we gather the features of interest from $t_l$ and ${v''_v}$, and then we use the gathered visual feature ${t'_l} \in \mathbb{R}{^{C \times 1}}$ acts as the query and the text embedding ${v'_l}$ as key and value to better collected semantic descriptions and output ${t_v}$. We definite $i\left( {1 < i < N} \right)$ as the current stage of the decoder. Thereafter, the query ${t_q^i}$ can be updated by ${t_v}$.
\begin{equation}
    t_q^{i + 1} = {f_{LN}}\left( {{f_{LN}}\left( {t_q^i + {t_v}} \right) + {f_{FFN}}({f_{LN}}(t_q^i + {t_v}))} \right)
\end{equation}
Where ${f_{LN}}( \cdot )$ and ${f_{FFN}}( \cdot )$ denote L2-normalization and a feed-forward network, respectively. Each decoder layer adds cross-modality information for visual-linguistic alignment.
\subsection{Training loss}
    In the training stage, we combine the loss function of OV-D with VLTVG~\cite{Yangetal-vltvg} to our proposed OV-VG framework. To encourage alignment between visual and language elements, we introduce and enhance the contrastive alignment loss~\cite{mdetr_contrastive_loss}. This loss ensures the text embedding and target object embedding are closer to each other than to other unrelated object embeddings. Specifically, we consider the text embedding as ${t_i}$, the number of proposal embeddings as ${N}$, and ${O_i^ + }$ represents the positive set of objects that align with ${t_i}$. The improved contrastive alignment loss supervises the degree of alignment between the text embedding and each proposal box to ensure that the output proposals are relevant to the sentence semantics, which is given by:
\begin{equation}
    {{\cal L}_{cts}} = \frac{1}{{\left| {O_i^ + } \right|}}\sum\limits_{j \in O_i^ + } { - \log \left( {\frac{{\exp \left( {t_i^T{o_j}/\tau } \right)}}{{\sum\nolimits_{k = 0}^{N - 1} {\exp \left( {t_i^T{o_k}/\tau } \right)} }}} \right)}
\end{equation}
where ${\tau}$ is a temperature parameter. The overall loss function 
which is as follows, 
\begin{equation}
    \mathcal{L} = \lambda_{\text{giou}}\mathcal{L}_{\text{giou}} + \lambda_{\text{L1}}\mathcal{L}_{\text{L1}} + \lambda_{\text{cts}}\mathcal{L}_{\text{cts}},
\end{equation}
where $\mathcal{L}_{\text{giou}}$, $\mathcal{L}_{\text{L1}}$, and $\mathcal{L}_{\text{cts}}$ denote the GIoU loss, L1 loss and contrastive alignment loss\cite{mdetr_contrastive_loss}, respectively. $\lambda_{\text{giou}}$, $\lambda_{\text{L1}}$ and $\lambda_{\text{cts}}$ are introduced to balance the above losses, we set the ${\lambda _{{\rm{L1}}}} = 5$ and ${\lambda _{{\rm{giou}}}} = {\lambda _{cts}} = 2$.

\begin{table*}[t]
\renewcommand{\arraystretch}{1.5}
\begin{center}
\caption{Compare with the methods of VG and OV-D structure framework}
\label{main_table}
\setlength{\tabcolsep}{2.0mm}{
\begin{tabular}{c|cc|c|c|cccc}
\hline
Method & Text Model & Vision Backbone & Pre-Training Data & Params(G) & Small & Middle & Large & Acc50 \\ \hline
 TransVG~\cite{Dengetal-TransVG} & BERT & ResNet50 & RefC & 149.5 & 0.0 & 0.04 & 7.17 & 2.57 \\
 VLTVG~\cite{Yangetal-vltvg} & BERT & ResNet50 & RefC & 151.3 & 0.0 & 0.04 & 8.05 & 2.78 \\
 VLTVG~\cite{Yangetal-vltvg} & CLIP & ResNet50 & RefC & 144.3 & 0.0 & 0.02 & 6.68 & 2.30 \\
 VLTVG~\cite{Yangetal-vltvg} & CLIP & CLIP & RefC & 144.4 & 0.0 & 0.02 & 7.97 & 2.74 \\
 Grounding DINO~\cite{liuetal-groundingDINO} & BERT & Swin-T & RefC & 172.5 & 0.0 & \textbf{0.08} & 7.07 & 2.59 \\
 Ours & CLIP & CLIP & RefC & 156.2 & \textbf{0.0} & 0.04 & \textbf{10.07} & \textbf{3.64} \\ \hline
 X-decoder~\cite{X-decoder} & CLIP & Focal-T & COCO,Cap4M,COCOK,RefCg & 39.3 & 0.0 & 13.39 & 14.73 & 13.32 \\
 X-decoder~\cite{X-decoder} & CLIP & Focal-L & COCO,Cap4M,COCOK,RefCg & 39.3 & 0.0 & 14.34 & 15.07 & 14.18 \\
 SEEM~\cite{seem} & CLIP & Focal-T & COCO,LVIS & / & 0.94 & 9.57 & 46.44 & 22.12 \\
 SEEM~\cite{seem} & CLIP & Focal-L & COCO,LVIS & / & 0.74 & 8.88 & 46.04 & 21.93 \\
 OpenSeeD~\cite{openseed} & CLIP & Swin-T & O365,COCO & / & 16.93 & 27.63 & 31.96 & 27.38 \\
 Kosmos-2~\cite{kosmos-2} & Kosmos2text & Kosmos2image & LAION-2B,COYO-700M & / & 0.75 & 18.33 & \textbf{62.679} & 30.70 \\
 Grounding DINO~\cite{liuetal-groundingDINO} & BERT & Swin-T & O365,GoldG,Cap4M & 172.5 & 7.48 & 34.63 & 53.88 & 37.38 \\
 Grounding DINO*~\cite{liuetal-groundingDINO} & BERT & Swin-T & O365,GoldG,Cap4M,RefC & 172.5 & \textbf{20.49} & 35.64 & 51.79 & 39.12 \\
 Ours & BERT & Swin-T & O365,GoldG,Cap4M,RefC & 173.1 & 18.15 & \textbf{38.80} & 55.27 & \textbf{41.55} \\ \hline
\end{tabular}}
\end{center}
\end{table*}

\subsection{Implementation Details}
To verify the performance of open-vocabulary visual grounding and prevent data leakage, we conduct experiments by training our models on the RefCOCO dataset\cite{yuetal-refcoco} and inference on the OV-VG dataset. For the image feature and text embedding extraction branches, we use ResNet50 and CLIP, respectively. We resize the images to 640 $\times$ 640 pixels and set the maximum text length to 256. Our experiments are conducted on two NVIDIA GeForce RTX 3090 GPUs using the AdamW optimizer with a learning rate of $1 \times {10^{-4}}$ and weight decay of $1 \times {10^{-5}}$. We utilize a batch size of 16 and train for ten epochs to facilitate a fair comparison with existing methods.\par

\begin{table*}[t]
\renewcommand{\arraystretch}{1.5}
\scalebox{1.0}{
\begin{minipage}{0.6\textwidth}
\begin{center}
\caption{Results with data leakage on our OV-VG dataset}
\label{Data_leakage_1}
\begin{tabular}{ccc|c|cccc}
\hline
Finetune & LGFA & TIQS & Pre-train Data & Small & Middle & Large & Acc50 \\ \hline
${\rm{ \times }}$ & ${\rm{ \times }}$ & ${\rm{ \times }}$ & RefC & 0.0 & 0.08 & 7.07 & 2.59 \\
${\rm{ \times }}$ & ${\rm{ \times }}$ & ${\rm{ \times }}$ & O365,GoldG,Cap4M & 7.48 & 34.63 & 53.88 & 37.38 \\
$\surd$ & ${\rm{ \times }}$ & ${\rm{ \times }}$ & O365,GoldG,Cap4M,RefC & \textbf{20.49} & 35.64 & 51.79 & 39.12 \\
$\surd$ & ${\rm{ \times }}$ & $\surd$ & O365,GoldG,Cap4M,RefC & 17.31 & 37.02 & 53.18 & 39.80 \\
$\surd$ & $\surd$ & ${\rm{ \times }}$ & O365,GoldG,Cap4M,RefC & 19.45 & 38.01 & 52.37 & 40.49 \\
$\surd$ & $\surd$ & $\surd$ & O365,GoldG,Cap4M,RefC & 18.15 & \textbf{38.80} & \textbf{55.27} & \textbf{41.55} \\
\hline
\end{tabular}
\end{center}
\end{minipage}
\begin{minipage}{0.45\textwidth}
\begin{center}
\caption{Ablation Study of finetune epoches}
\label{finetune}
\begin{tabular}{c|cccc}
\hline
Epoch & Small & Middle & Large & Acc50 \\ \hline
1 & 18.15 & 38.80 & \textbf{55.27} & \textbf{41.55} \\
2 & \textbf{18.74} & \textbf{39.12} & 53.85 & 41.31 \\
3 & 13.73 & 29.08 & 34.10 & 28.53 \\
4 & 18.09 & 37.45 & 44.34 & 36.97 \\
5 & 16.85 & 38.48 & 46.54 & 38.05 \\
6 & 12.17 & 31.86 & 38.58 & 31.25 \\
\hline
\end{tabular}
\end{center}
\end{minipage}
}
\end{table*}
\section{Experiments} 
  \label{experiments}
To verify the effectiveness of our method, we compare it with existing state-of-the-art (SOTA) methods both on regular VG and open-vocabulary frameworks, as shown in Table~\ref{main_table}. The top part compares regular VG method without data leakage, and the bottom part is the existing open-vocabulary method with data leakage. In regular VG, we select TransVG and VLTVG as representatives of VG framework, and we also employ Grounding DINO without pre-training as an open-vocabulary structure. Unlike traditional VG evaluation settings, we do not perform experiments on RefCOCO, RefCOCO+, and RefCOCOg datasets. This decision is based on the fact that, for open-vocabulary problems, the training set of the aforementioned three datasets can be considered identical, as they all consist of base classes. Instead, we exclusively evaluate our method on the OV-VG dataset containing only novel categories. \par
To provide a more detailed analysis of the results, we take the size variation of targets about our OV-VG dataset into consideration. We categorized the target sizes based on their bounding boxes into large (box size larger than 96 $\times$ 96), middle (in the middle of 32 $\times$ 32 and 96 $\times$ 96), and small (smaller than 32 $\times$ 32) refer to object detection, each contains 1537, 4868 and 3595 images, respectively. \par
As shown at the top of Table~\ref{main_table}, the first two rows show the results of the original TransVG and VLTVG. Since the DETR in VLTVG has been pre-trained with BERT, the Acc50 performance revealed in our OV-VG dataset is 2.78$\%$. After replacing BERT and DETR with CLIP visual and text backbone, the Acc50 remains almost unchanged. However, when we change the text part of VLTVG to CLIP, and the visual retains DETR, the performance declines by 0.48$\%$. We choose Grounding DINO as the open-vocabulary method and do not pre-train it on a large amount of data, and the Acc50 is worse than the VLTVG framework. Our method outperforms both regular VG and open-vocabulary framework, and achieves 3.64$\%$ average Acc50 and 10.07$\%$ Acc50 on large targets.\par
As shown at the bottom of Table~\ref{main_table}, we compare our method with existing open-vocabulary frameworks with data leakage. X-decoder, SEEM and Kosmos-2 seem unable to solve the small target problem of VG, resulting in almost zero Acc50. Since SEEM has been pre-trained on LVIS, where our data annotation comes from. Data in the same domain will bring performance improvements. The performance of OpenSeeD is much higher than X-decoder and SEEM, especially on small targets. Kosmos-2 outperforms on large targets. Since Grounding DINO can understand long sentences better, it performs best. Grounding DINO* means the finetuning result on RefCOCO, it can better detect the small targets. After that, we add our modules on Grounding DINO, which achieves the SOTA results.\par

\begin{table}[t]
\renewcommand{\arraystretch}{1.5}
\begin{center}
\caption{Ablation study of our proposed network on OV-VG datasets}
\label{Ablation_study}
\setlength{\tabcolsep}{1.5mm}{
\begin{tabular}{cc|cccc}
\hline
LGFA & TIQS & Small & Middle & Large  & Acc50 \\ \hline
${\rm{ \times }}$ & ${\rm{ \times }}$ & 0.0 & 0.04 & 9.10 & 3.29 \\
$\surd$ & ${\rm{ \times }}$ & 0.0 & 0.06 & 9.74 & 3.53 \\
${\rm{ \times }}$ & $\surd$ & 0.0 & \textbf{0.08} & 9.29 & 3.38 \\
$\surd$ & $\surd$ & 0.0 & 0.04 & \textbf{10.07} & \textbf{3.64} \\
\hline
\end{tabular}}
\end{center}
\end{table}
\begin{table}
\renewcommand{\arraystretch}{1.5}
\begin{center}
\caption{Results with data leakage on our OV-PL dataset}
\label{Data_leakage_2}
\setlength{\tabcolsep}{1.2mm}{
\begin{tabular}{c|cc|ccc}
\hline
Method & Pre-train Data & Category & R@1 & R@5 & R@10 \\ \hline
GLIP~\cite{lietal-GLIP} & \multirow{2}{*}{O365,GoldG,Cap4M} & Base & 64.5 & 77.1 & 79.7 \\
 &  & Base+Novel & 41.6 & 56.0 & 60.2 \\ \hline
FIBER~\cite{douetal-FIBER} & \multirow{2}{*}{\begin{tabular}[c]{@{}c@{}}COCO,SBU,GCC,ViGe\\ O365,GoldG,Flickr30k\end{tabular}} & Base & 76.9 & 83.5 & 84.0 \\
 &  & Base+Novel & 59.7 & 70.6 & 72.7 \\
\hline
\end{tabular}}
\end{center}
\end{table}
\subsection{Ablation Study}
In this subsection, we conduct the ablation studies on our OV-VG dataset. Table~\ref{Ablation_study} presents the effectiveness of each component in the proposed method on our OV-VG dataset. Numerically, LGFA improves 0.22$\%$ and TIQS improves 0.09$\%$ in average Acc50. At the same time, LGFA and TIQS improve 0.64$\%$ and 0.19$\%$ in large targets, respectively. Although the overall improvement is insignificant, it is still considerable for such a low overall accuracy. \par
To further verify the effectiveness of our proposed method, we add the LGFA and TIQS in Grounding DINO. After finetuning the model for one epoch, we report the numerical results as shown in Table~\ref{Data_leakage_1}. When we only add TIQS and further finetune Grounding DINO, Acc50 improves by 0.68$\%$, while LGFA improves by 1.37$\%$. Adding both LGFA and TIQS, the Acc50 improves 2.43$\%$ and the Acc50 in small, middle, and large targets for 0.84 $\%$, 1.78 $\%$ and 2.19 $\%$, which also proves the effectiveness of our proposed components.\par
\subsection{Data Leakage}
In this subsection, we analyze the data leakage of existing open-vocabulary methods and present the results of our OV-VG dataset on Grounding DINO and our OV-PL dataset on GLIP.\par
The currently released Grounding DINO has two versions: one is pre-trained on Object365, GoldG(GoldG is a subset of GoldG+ excluding COCO images, GoldG+ containing 1.3M data including Flickr30k, VG caption and GQA) and Cap4M dataset, another is pre-trained on Object365, GoldG, OpenImage, Cap4M, COCO and RefCOCO. Since the latter uses COCO for training, images of OV-VG may have been seen, so we chose the former version to illustrate the data leakage. In summary, Grounding DINO utilizes a large amount of data for training, which means that the novel categories in our OV-VG dataset have been leaked out (the Grounding DINO has seen the novel categories during training).\par
\begin{table*}[t]
\renewcommand{\arraystretch}{1.2}
\begin{center}
\caption{Results on OV-VG 100 novel categories}
\label{Classes}
\setlength{\tabcolsep}{2.0mm}{
\begin{tabular}{cccccccccc}
\hline
air conditioner & antenna & apron & awning & baby buggy & banner & bath towel & belt & blanket & bracelet \\
67 & 21 & 55 & 42 & 70 & 37 & 38 & 52 & 48 & 37 \\ \hline
bucket & button & cabinet & camera & candle & Christmas tree & clock tower & coat & cone & crossbar \\
58 & 15 & 28 & 42 & 32 & 51 & 80 & 32 & 44 & 2 \\ \hline
curtain & cushion & dog collar & doorknob & drawer & dress & earring & faucet & flag & glove \\
47 & 19 & 22 & 27 & 17 & 51 & 11 & 51 & 37 & 35 \\ \hline
goggles & handle & hat & headlight & helmet & hinge & home plate & jacket & jean & jersey \\
58 & 18 & 48 & 17 & 69 & 10 & 52 & 42 & 34 & 39 \\ \hline
lamp & lamppost & license plate & lightbulb & mirror & napkin & necklace & painting & pillow & pipe \\
38 & 28 & 34 & 19 & 37 & 32 & 41 & 30 & 29 & 22 \\ \hline
place mat & plastic bag & plate & pole & pot & reflector & saddle & shirt & shoe & short pants \\
46 & 54 & 46 & 23 & 43 & 18 & 8 & 42 & 41 & 56 \\ \hline
signboard & skirt & ski boot & ski pole & soap & sock & speaker & spectacles & statue & stove \\
29 & 58 & 33 & 6 & 26 & 33 & 43 & 70 & 51 & 57 \\ \hline
streetlight & street sign & sunglasses & sweater & sweatshirt & tablecloth & taillight & tarp & toilet tissue & towel \\
26 & 28 & 50 & 58 & 48 & 45 & 35 & 67 & 45 & 36 \\ \hline
toy & trash can & tray & trousers & vent & wall socket & watch & wet suit & wheel & wristlet \\
41 & 61 & 54 & 50 & 36 & 37 & 52 & 70 & 20 & 60 \\ \hline
balloon & basket & bathtub & blender & blouse & bun & butter & calendar & chandelier & windshield wiper \\
67 & 61 & 71 & 80 & 48 & 57 & 16 & 13 & 81 & 11 \\
\hline
\end{tabular}}
\end{center}
\end{table*}
\begin{figure}
\begin{center}
   \includegraphics[width=1.0\linewidth]{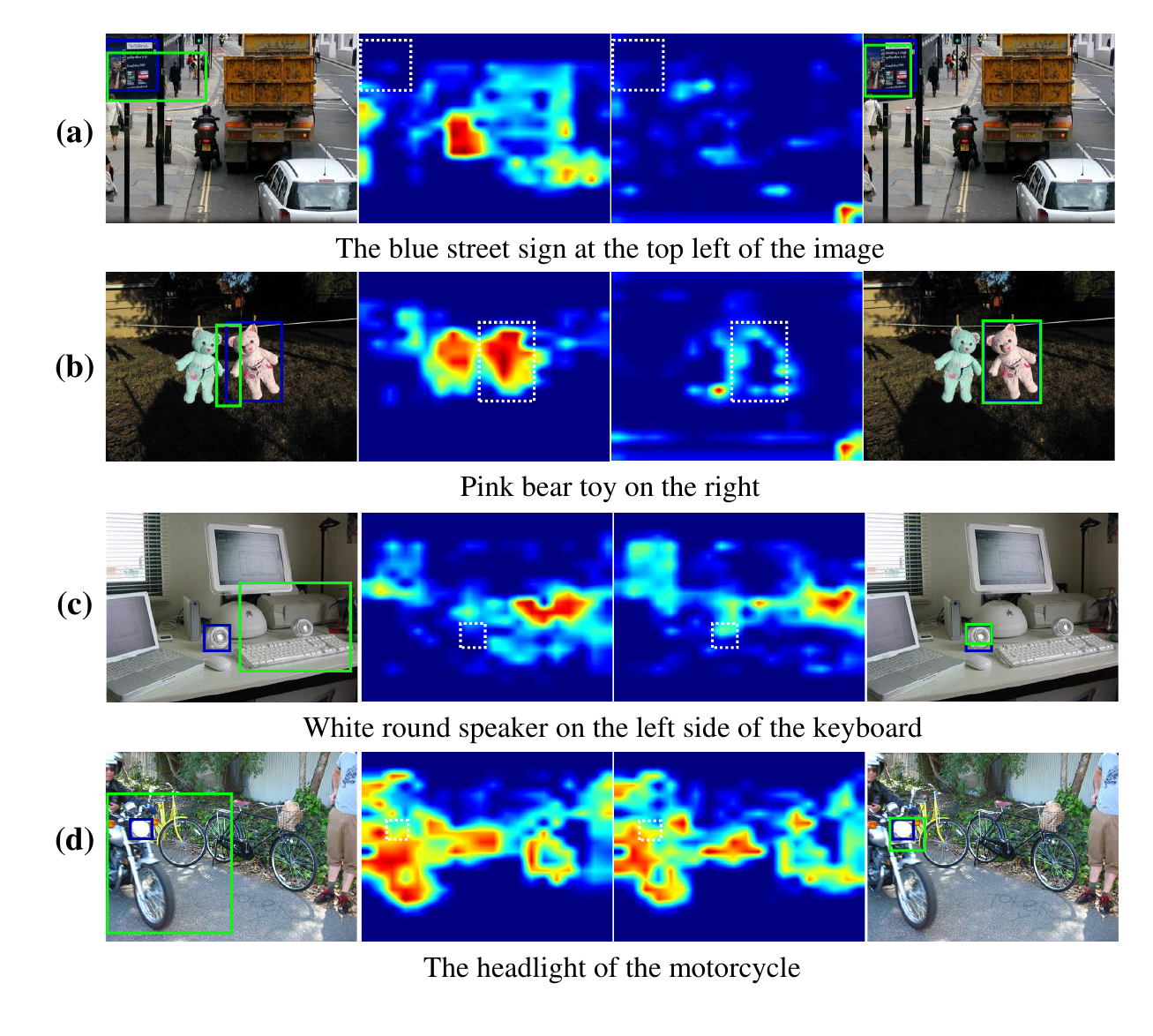}
\end{center}
   \caption{Visualization results. Left two columns mean the Grounding DINO and right two columns indicate our method. (a) and (b) are regular open-set results, (c) and (d) are results with data leakage. White dashed boxes on the feature map represent the ground truth.}
\label{visualization_fusion}
\end{figure}
\begin{figure*}
\begin{center}
   \includegraphics[width=1.0\linewidth]{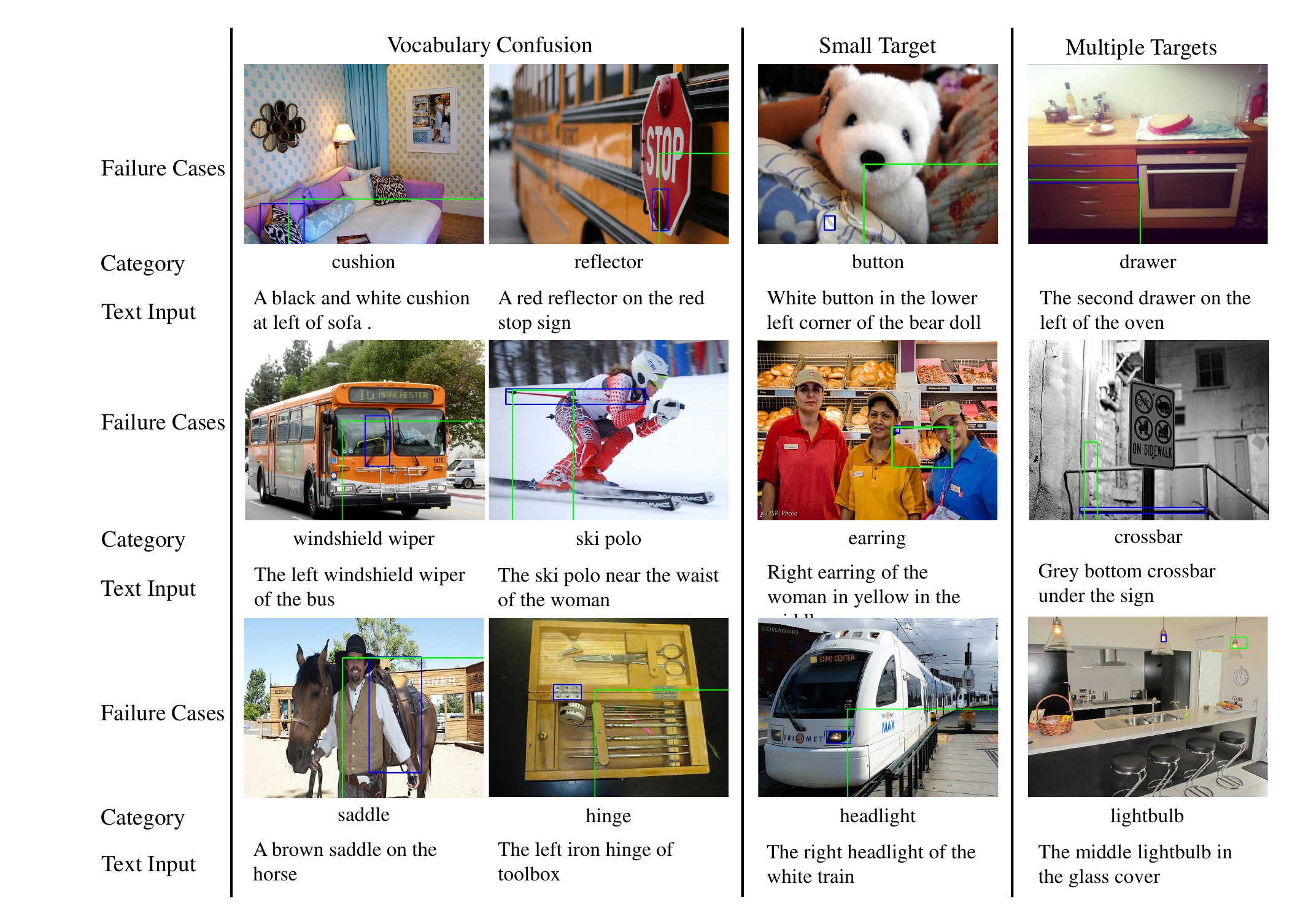}
\end{center}
   \caption{Failure cases results of our OV-VG dataset. Blue represents ground truths and green means predict boxes. The first row denotes predict result, the second row is category name and the third row is input text. The left two columns are vocabulary confusion results, the third column is small target examples and the right most column is multiple target problem.}
\label{failure_cases}
\end{figure*}
Firstly, we test our OV-VG dataset by the original Grounding DINO directly. Then we finetune it with RefCOCO training set. Finally, we add our proposed LGFA and TIQS in Ground DINO to verify the validity of our proposed models in the case of data leakage. Table~\ref{Data_leakage_1} certificates that after pre-trained in Object365, GoldG, and Cap4M, the performance substantially increased due to the data leakage. The results of Grounding DINO finetuned on RefCOCO after incorporating the LGFA and TIQS modules are shown in Table~\ref{finetune}. As the number of finetune epochs increases, the original Grounding DINO model tends to forget the data leakage information (including novel categories learned in pre-trained dataset) previously and gradually converges to the base categories in RefCOCO.\par
Table~\ref{Data_leakage_2} presents the results of existing phrase localization foundation models (GLIP and FIBER) on our OV-PL dataset. As we can see, training on Object365, GoldG, and Cap4M, GLIP achieves remarkable performance, 64.5$\%$ Recall@1 on base categories and 41.6$\%$ Recall@1 on base and novel categories. At the same time, by increasing the pre-training data on COCO, SUB Captions Conceptual Captions, Visual Genome, and Flickr30k, FIBER achieves a Recall@1 of 76.9$\%$ on base categories, which is 12.4$\%$ higher than GLIP, and a Recall@1 of 59.7$\%$ on both base and novel categories, which is 18.1$\%$ higher than GLIP. Nowadays, most researchers focus on foundation modules that utilize large amounts of data pre-training, and fewer researchers pay attention to the data leakage problems, which means performance improvement is likely 
data leakage during training.

\subsection{Dataset Analysis and Failure Cases}
To better analyze the characteristics and challenges of our OV-VG dataset, we report the Acc50 of 100 novel categories, as shown in Table~\ref{Classes}. Common categories with not too small sizes can be well detected (Acc50 is greater than or equal to 70$\%$), such as 'baby buggy', 'clock tower', 'spectacles', 'wet suit', 'blender', 'chandelier' and 'bathtub'. However, several categories are almost completely undetectable (Acc50 is less than or equal to 20$\%$), such as 'crossbar', 'button', 'drawer', 'earring', 'hinge', 'saddle', 'ski pole' and 'windshield wiper'. We classified these categories of failure cases through visual analysis, as shown in Fig.~\ref{failure_cases}. These failure categories can be classified into three parts: (1) vocabulary confusion. Detectors can not recognize objects represented by such complex vocabulary, which will lead to predicting completely non-corresponding boxes, such as 'The left windshield wiper of the bus' and 'The left iron hinge of toolbox'. (2) Small Target. Some categories are too small to detect in the image, such as 'button' and 'earring', especially when the image scene is more complex. (3) Multiple Targets. There are multiple objects in the image, and we have to use more precise orientation information when describing them, such as 'the second drawer' and 'the middle lightbulb'. In summary, the above three challenges are the difficulties of the OV-VG dataset, especially when the corresponding target contains two or three challenges simultaneously, such as 'The second button in the shirt'. Through experiments and visual analysis, our OV-VG is challenging in not only the task but also the dataset.

\begin{figure*}
    \centering
    \begin{minipage}[t]{1.0\linewidth}
        \centering
        \includegraphics[width=1.0\linewidth]{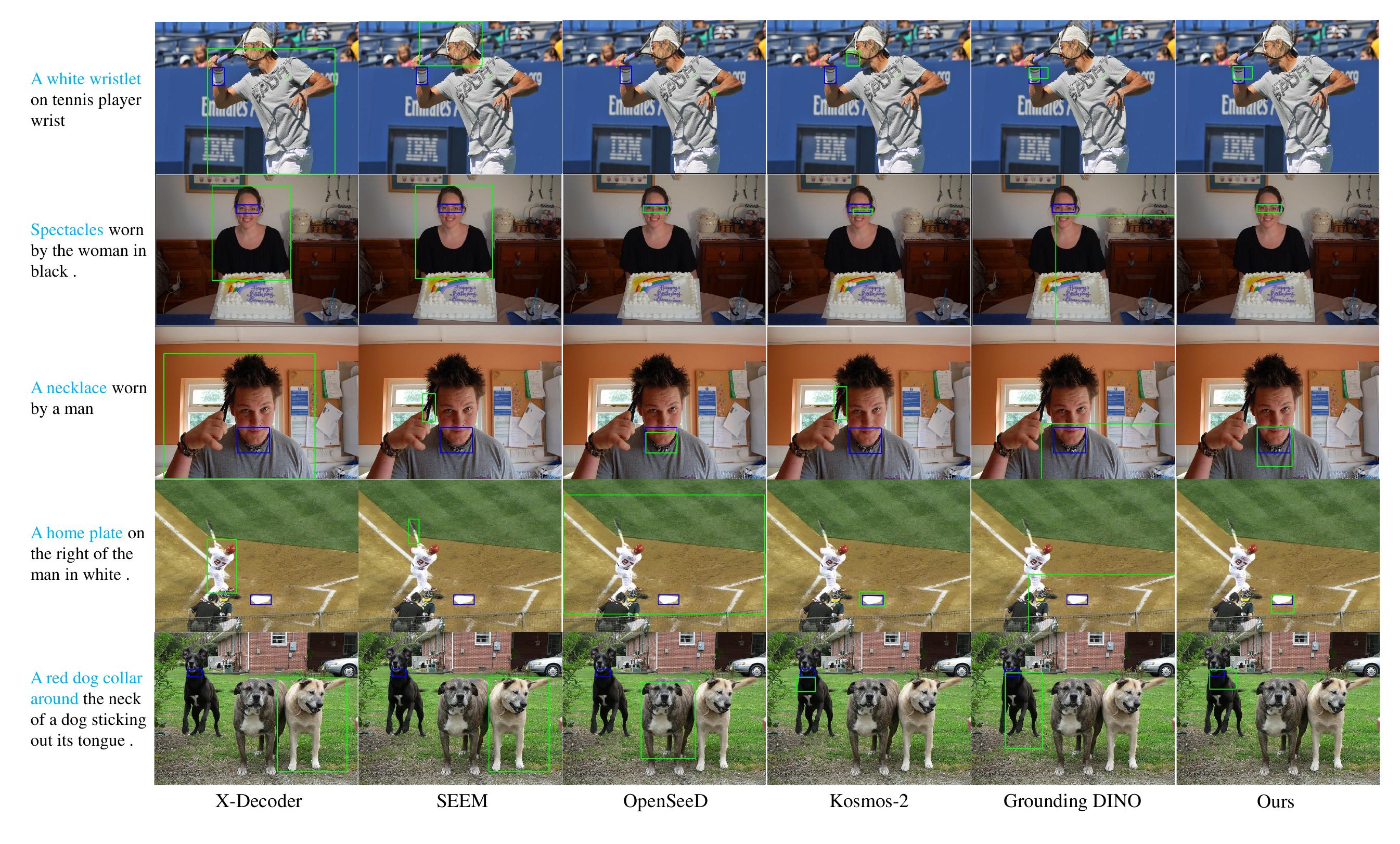}
        \caption{Visualization results of existing open-vocabulary methods and ours. The first column means input sentence, the last six columns denote the open-vocabulary methods and ours. Blue represents ground truths and green means predict boxes.}
        \label{compare}
    \end{minipage}
    \begin{minipage}[t]{1.0\linewidth}
        \centering
        \includegraphics[width=1.0\linewidth]{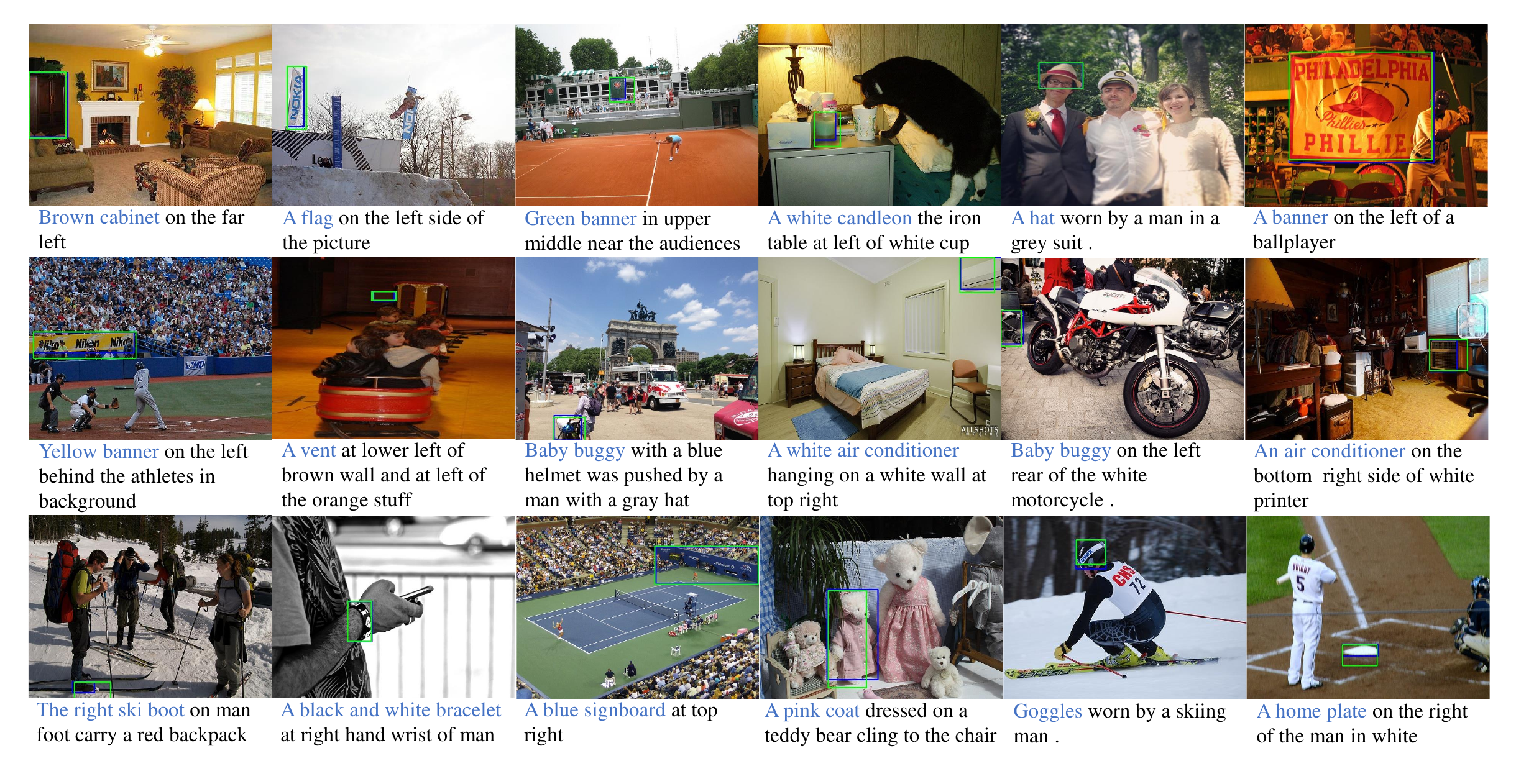}
        \caption{Visualization results of the predict boxes. Novel categories are indicated in blue font in the input sentence.}
        \label{visualization_boxes}
    \end{minipage}
 \end{figure*}

\subsection{Visualization Experiments}
In Fig.~\ref{visualization_fusion} (a) and (b), we visualize the Grounding DINO (the left two columns) and our proposed method (the right two columns) on the OV-VG dataset. Grounding DINO approach often exhibits a tendency to detect all objects indiscriminately. Unfortunately, language cues are frequently overlooked during such instances, leading to inaccuracies in the detection process. Particularly evident in Figure~\ref{visualization_fusion}(b), when confronted with two identical novel category targets, the Grounding DINO method often struggles to provide the precise target bounding box. In contrast, our approach effectively leverages textual information to identify and localize the correct target box accurately.
\par Existing VG datasets generally contain few small objects, which poses a significant challenge for visual-linguistic alignment when the target object is diminutive. As exemplified in Figure~\ref{visualization_fusion} (c) and (d), we illustrate the outcomes of small target visual grounding in scenarios involving data leakage within our OV-VG dataset. When dealing with a small object from a novel category, Grounding DINO often misinterprets the textual input, leading to the prediction of 'keyboard' and 'motorcycle' instead of the actual 'speaker' and 'headlight.' Notably, our method excels in addressing both open-vocabulary challenges and data leakage situations. \par

We perform a comparative analysis of the visualization between the existing open-vocabulary methods and ours. As for X-Decoder and SEEM, we use the smallest enclosing rectangle of the segmentation result as the box. At the same time, we choose the top-1 results of X-Decoder, SEEM, OpenSeeD, Kosmos-2, and Grounding DINO. As shown in Fig~\ref{compare}, X-Decoder~\cite{X-decoder} tends to predict the base category in a sentence, such as 'woman', 'man', and 'dog'. SEEM~\cite{seem} tries to understand the sentence and image information, such as 'knife' and 'baseball bat'. OpenSeeD~\cite{openseed} can better understand the sentence and image. However, mistakes can also be made when encountering confusing novel vocabulary, such as 'dog collar'. Kosmos-2~\cite{kosmos-2} can effectively handle large objects, but its ability to handle small objects is much weaker. Grounding DINO~\cite{liuetal-groundingDINO} can identify novel categories, but the positioning is not accurate. Our method can better achieve visual-linguistic alignment and better predict the target object.\par
Fig.~\ref{visualization_boxes} shows the visualization results of our method on the OV-VG dataset. It can be observed that when we input a long sentence about the novel category, the model can accurately locate the described target, regardless of the complexity of the image or the length of the target description, such as predicting the 'Baby buggy' in the sentence 'Baby buggy with a blue helmet was pushed by a man with a gray hat'.
\par

\section{Conclusion} 
\label{conclusion}
In this paper, we comprehensively explore problem settings in the context of open-vocabulary visual grounding and open-vocabulary phrase localization. To facilitate research in this area, we introduce two novel benchmark datasets. First, we provide insights into the dataset structures and offer a detailed analysis of the underlying objectives for these two tasks. Subsequently, we establish a solid foundation by presenting state-of-the-art baselines for OV-VG and OV-PL datasets. To advance the field, we propose a novel OV-VG framework incorporating LGFA and TIQS modules to enhance visual-linguistic comprehension. We rigorously evaluate our method through extensive experiments on the OV-VG dataset, considering potential data leakage scenarios. Additionally, we delve into the complexities and obstacles presented by the OV-VG dataset by introducing 100 novel categories, shedding light on its challenges. Furthermore, we compare our approach with existing SOTA open-vocabulary methods and thoroughly analyze the results, demonstrating the inherent difficulty and significance of the OV-VG task. We also validate the rationality of our methodology through visual experiments.
\par Given the suboptimal performance of existing methods when data leakage is absent, our future research direction focuses on broadening the representation of novel categories and devising a more elegant pipeline to address these issues effectively.

{\small
\bibliographystyle{IEEEtran}
\bibliography{refbib}
}

\end{document}